\documentclass{article}
\usepackage{spconf,amsmath,graphicx}
\usepackage{amssymb}
\usepackage{algorithm}
\usepackage{algorithmic}
\usepackage{comment}
\usepackage{url}
\usepackage{amsmath,graphicx}
\usepackage[noadjust]{cite}
\usepackage{array}
\usepackage{gensymb}
\usepackage[utf8]{inputenc}
\usepackage[T1]{fontenc}

\title{Scene Completeness-Aware Lidar Depth Completion for Driving Scenario}
\name{Cho-Ying Wu  \hspace*{ 0.1 in} Ulrich Neumann}
\address{University of Southern California}
\begin{document}
%
\maketitle
\begin{abstract}
This paper introduces Scene Completeness-Aware Depth Completion (SCADC) to complete raw lidar scans into dense depth maps with fine and complete scene structures. Recent sparse depth completion for lidars only focuses on the lower scenes and produces irregular estimations on the upper because existing datasets, such as KITTI, do not provide groundtruth for upper areas. These areas are considered less important since they are usually sky or trees of less scene understanding interest. However, we argue that in several driving scenarios such as large trucks or cars with loads, objects could extend to the upper parts of scenes. Thus depth maps with structured upper scene estimation are important for RGBD algorithms. SCADC adopts stereo images that produce disparities with better scene completeness but are generally less precise than lidars, to help sparse lidar depth completion. To our knowledge, we are the first to focus on scene completeness of sparse depth completion. We validate our SCADC on both depth estimate precision and scene-completeness on KITTI. Moreover, we experiment on less-explored outdoor RGBD semantic segmentation with scene completeness-aware D-input to validate our method.

\end{abstract}

\section{Introduction}
\label{sec:intro}
Depth is widely used to represent scene geometry~\cite{wu2023meta,wu2020geometry,wu2022toward,wu2023inspacetype,xu2020grid}.
Autonomous driving usually adopts lidars as the main depth acquisition sensors due to their high precision and practicability on outdoor depth sensing. However, lidar scans are limited to number of scanlines and spatial resolutions, and thus they are sparse when aligned with images. Recent research on lidar depth completion for autonomous driving tries to complete sparse lidar depth into a dense map \cite{uhrig2017sparsity, Ma2017SparseToDense,qiu2018deeplidar,ma2019self,zhong2019deep,leedepth,wang2018pnp,imran2019depth, chodosh18, chen2019learning, shivakumar2019dfusenet} using KITTI Depth Completion Dataset \cite{geiger2012we}. 

However, their depth map processing and evaluations always crop out the upper side of maps for two reasons. First, these upper side areas are usually sky or trees of low scene understanding interest. Second, lidars are active sensors with limited scanlines and smaller vertical field-of-view than cameras. Thus, most lidar scans do not span the whole image height and are concentrated on the lower parts of images. For KITTI, topside 1/3 to 1/4 areas are unscanned by lidars. Also, KITTI's depth groundtruth is acquired by accumulating 3D point clouds with a 64-scanline lidar. Hence their groundtruth are also concentrated on the lower parts of images. Both of KITTI's quantitative and qualitative evaluations focus only on the lower parts.
\begin{figure}[bt!]
    \centering
    \includegraphics[width=1.0\linewidth]{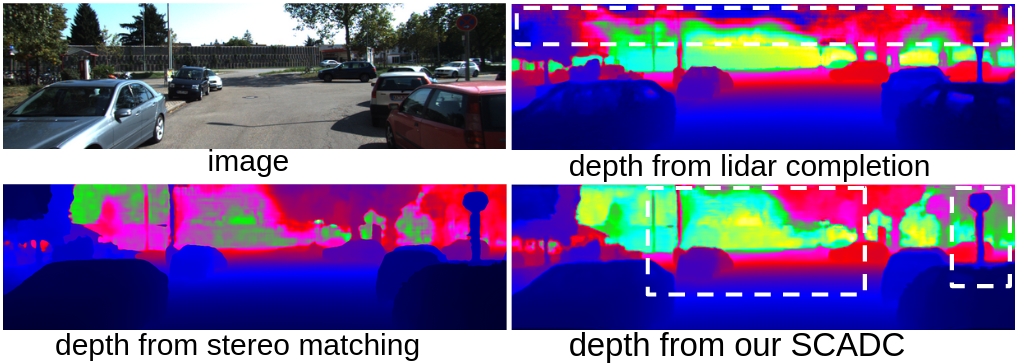}
    \vspace{-20pt}
    \caption{\textbf{Comparison of depth from stereo matching network \cite{chang2018pyramid}, depth completion network \cite{ma2019self}, and our SCADC.} Our results leverage advantages of both stereo matching, which have more \textit{structured} upper scenes, and lidars, which have more \textit{precise} depth measurements.}
    \label{teaser}
    \vspace{-15pt}
\end{figure}

Nevertheless, upper scenes are especially important under several autonomous driving scenarios, such as a huge truck beside or just in front occupies a large area of the upper scene when close enough. Traffic signs or lights are important road structures extending to the upper parts. Although more and more research focuses on multi-modal learning from images and depth, \textit{scene incompleteness issue is mostly ignored} for the following reasons. First, depth completion is treated as a \textit{standalone task} in previous works without validating their completed depth maps on other scene understanding tasks such as semantic segmentation. Second, not enough data of large objects extending to the upper scenes are collected, and thus the issue is generally omitted. However, autonomous driving needs to take care of all kinds of scenarios to prevent accidents and thus needs more attention to scene completeness issue.

In contrast, stereo matching produces disparity maps with much structured upper scenes because disparities are derived from images. However, stereo matching is known for less reliable depth measurements for far-range sensing and edge bleeding artifact \cite{wang2014stereo}, which produces distorted shapes.

In this work, we leverage the completeness of disparity maps to help sparse depth completion. To our knowledge, we are the first one focusing on the scene completeness issue of depth completion. Our Scene Completeness-Aware Depth Completion (SCADC) fuses depth estimations from a stereo matching network and a lidar depth completion network. We propose Attentional Point Confidence (APC) to regress confidence maps to fuse multi-modal information. Later, we use a stacked hourglass network to refine estimations stage by stage with groundtruth. Output examples are in Fig. \ref{teaser}. We use both quantitative and qualitative comparisons to validate that our SCADC combines the advantages of stereo cameras and lidars, producing both scene completeness-aware and precise depth maps.

Unlike previous works that treat depth completion as a standalone task, we further validate our scene completeness-aware depth on semantic segmentation. We use SSMA \cite{valada19ijcv}, a high-performing outdoor RGB-D semantic segmentation framework, as the baseline to show that our recovered depth could help better scene understanding.

\section{Related Work}
\label{sec:related}

\textbf{Sparse Depth Completion.} Recent work of sparse depth completion works on completing lidar depth using real-world data from KITTI Depth Completion Benchmark. Sparse-to-Dense \cite{Ma2017SparseToDense} stacks sparse depth maps and images to form a 4-channel input to a ResNet-based depth completion network. SSDC \cite{ma2019self} uses ego-pose coherence with a photometric loss to regress depth. CSPN \cite{cheng2018learning} and Non-local \cite{park2020non} adopt convolutional spatial propagation to enhance local information. Other studies, such as Deep-Lidar \cite{qiu2018deeplidar} and PwP \cite{xu2019depth} adopt an extra surface normal regression to help the depth estimation. However, these methods either crop out the upper scene of depth maps or produce random structures on the upper areas since KITTI only provides groundtruth for lower scenes. Besides, they work on depth completion alone without further vision applications using their depth maps. By contrast, we utilize scene completeness of disparity maps to help depth completion. In addition, we use our completed depth maps to help outdoor RGBD semantic segmentation, which further shows the values of depth completion.  

\begin{figure*}[hbt!]
    \centering
    \includegraphics[width=1.0\linewidth]{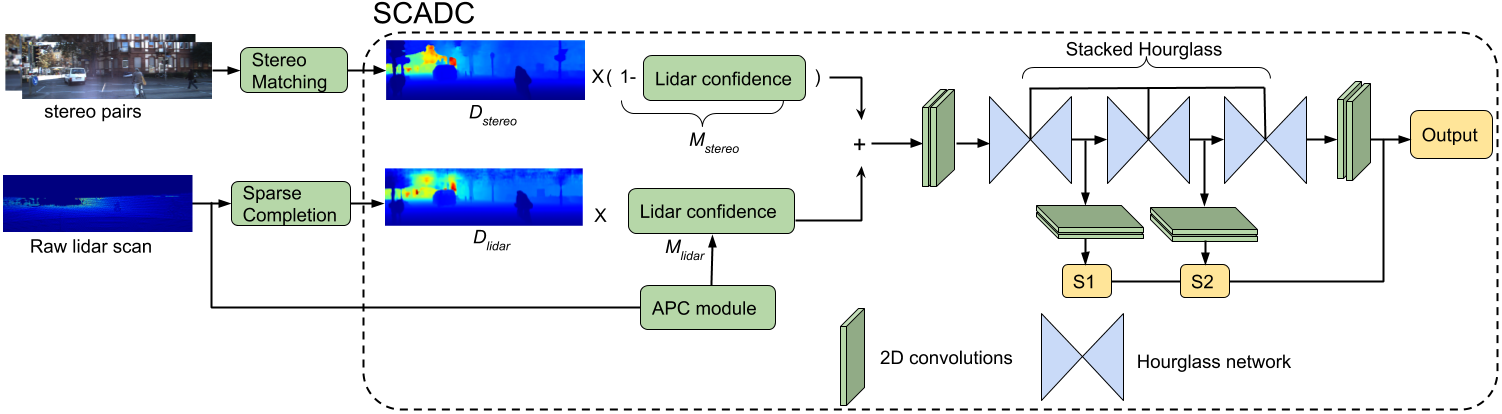}
    \vspace{-23pt}
    \caption{Network pipeline of our SCADC.}
    \label{pipeline}
    \vspace{-15pt}
\end{figure*}
\begin{figure}[bt!]
    \centering
    \includegraphics[width=1.0\linewidth]{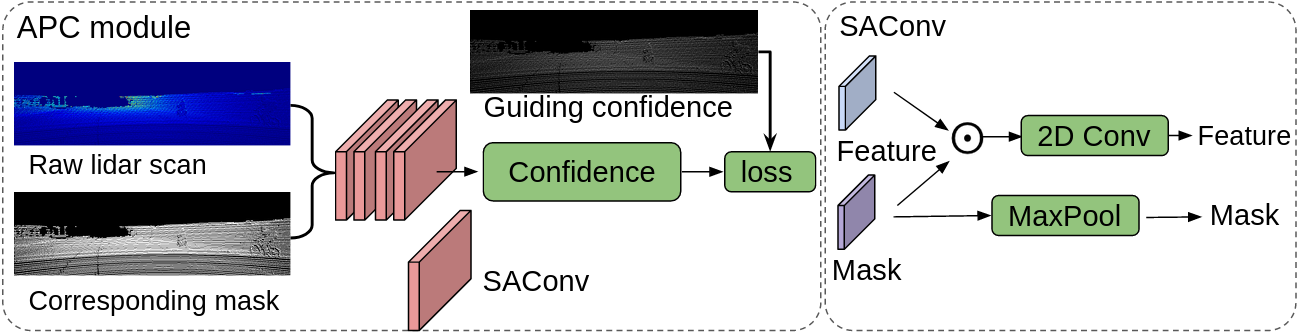}
    \vspace{-22pt}
    \caption{\textbf{Structure of APC module and Sparisty Attentional Convolution (SAConv)} \cite{zhong2019deep}. $\odot$ is for point-wise product.}
    \label{APC_SACONV}
    \vspace{-15pt}
\end{figure}

\textbf{Stereo Matching.} Stereo matching is a fundamental problem in computer vision. Traditional SGM and variants \cite{hirschmuller2007stereo,hirschmuller2005accurate,gehrig2009real,hermann2012iterative} match left/right frame features and output sparse disparity estimations. Recent stereo matching methods \cite{chang2018pyramid, Guo_2019_CVPR, Zhang_2019_CVPR} using neural networks could estimate dense disparity maps. They usually aggregate features into a cost volume and then apply 3D convolutions to regress disparities. The estimated dense disparities have more structured upper scenes than lidar completion since they are directly derived from image space with upper scene details. However, stereo matching usually suffers from edge bleeding that estimated disparities bleed out from object contours and form distorted areas \cite{forsyth2003modern,wang2014stereo}. Further, stereo matching-based methods are unreliable for long-range sensing or areas without textures, and in general they produce depth less precise than lidar measurements. Our SCADC combines advantages of preciseness of lidars at lower scenes and structured upper scenes of stereo matching to better utilize the two modalities.

\textbf{RGBD Semantic Segmentation.} There are much fewer works on RGBD outdoor semantic segmentation than indoors since high-quality outdoor depth maps are hard to get to match information from image and depth domains. SSMA \cite{valada19ijcv} combines two AdapNet++ \cite{valada19ijcv} branches and densely fuses information from images and depth encoders with a decoder to regress the depth map. We adopt SSMA and validate completed depth from our SCADC on outdoor semantic segmentation.

\section{Methods}
\label{sec:methods}

The whole network design of our SCADC is in Fig. \ref{pipeline}. Our goal is to construct a network for sensor fusion, which takes advantage of depth from stereo matching with more structured upper scenes, and depth from lidar completion with higher precision, to generate both scene completeness-aware and precise depth maps.

PSMNet \cite{chang2018pyramid} and SSDC \cite{ma2019self} are adopted as our base methods for stereo matching and lidar completion respectively. We use the estimated depth maps from two modalities, $D_{stereo}$ and $D_{lidar}$, as inputs to our SCADC. SCADC consists of two parts, multi-modal fusion and regression with a stacked hourglass network.

At the multi-modal fusion stage, we utilize the early fusion strategy. Early fusion incorporates multi-modal information before an encoder stage and has the advantage of retaining finer local structures and neighborhood relationships. Opposed to early fusion, late fusion is usually adopted for multi-modal learning with modalities from different domains to capture higher-level semantics, such as fusing information of images and depth \cite{zhong2019deep, Jaritz_2018}. Our SCADC operates information fusion only on the \textit{depth} domain, and thus early fusion of retaining local features and structures is more desirable.

We propose a novel confidence regression module, Attentional Point Confidence (APC), to estimate the pixel-level confidence of lidars, $M_{lidar}\in [0,1]^{H\times W}$, where $H$ and $W$ are height and width of inputs. APC decides for each pixel which modality is \textit{more probable} to estimate \textit{more reliable depth}. Previous works \cite{qiu2018deeplidar, van2019sparse} also use confidence maps for RGBD fusion without direct supervision on confidence regression. However, for stereo cameras/lidars fusion, since we have priors that depth from stereo matching is more structured on upper scenes and depth from lidar scans is more precise in general, using direct supervision on confidence regression could make the network generate better confidence maps of maintaining and combining both advantages from stereo cameras and lidars.

We create guiding confidence $M_g$ from the raw lidar scans. Lidar measurements are comparatively precise, so pixel positions at raw lidar points should have higher confidence. We set their scores to 1. Next, the depths of neighboring pixels are generally similar, so we dilate confidence at each raw lidar point using Gaussian kernels to obtain $M_g$. Kernel size and variance are based on point density of raw lidar scans. We find density along a scanline for KITTI is 44.6\% at the center and 30.6\% near the left/right side. Thus, we use a $3\times 3$ kernel and choose a variance which makes confidence scores drop to half with a 1-pixel distance from the center. $M_g$ provides better priors of confidence maps for learning. Note that (1) $M_g$ is not a hard constraint. The network could still generate confidence maps learned from all loss combinations. (2) Estimating a confidence map from one modality is enough for the probabilistic fusion of two modalities using the sum to 1 constraint. We choose to estimate lidars' since their scanned points are generally more precise, giving convenience to create $M_g$. 

Sparse data are intrinsically hard for CNN to extract effective features. In APC, We utilize Sparsity-Attentional Convolution (SAConv) \cite{zhong2019deep} to extract features from sparse lidar maps. SAConv attends on feature extraction of each nonzero point with an extra mask to keep track of visibility. After regressing $M_{lidar}$, we calculate the confidence loss as $L_c = \|M_{lidar}-M_g\|^2_2$. Structures are illustrated in Fig. \ref{APC_SACONV}. The confidence for stereo is $M_{stereo}=1-M_{lidar}$, and the fused depth is $ D_{f} = D_{stereo}\times M_{stereo}+ D_{lidar}\times M_{lidar}.$


The second stage is depth regression. We use a stacked hourglass network \cite{newell2016stacked} with dense connections for regressing depth. Our stacked hourglass network consists of 3 cascaded encoder-decoder structures. It has the advantage of refining depth maps stage by stage when compared with mostly used single encoder-decoder of FCN-like structure in other depth completion works \cite{ma2019self} \cite{Ma2017SparseToDense} \cite{zhong2019deep} \cite{Jaritz_2018}. The stacked hourglass produces 3 stage outputs ($S1$, $S2$, and $S3$). We further use skip connection and densely connect each corresponding layer of these hourglasses and feed the regressed depth to every subsequent stage to enhance information flow. Finer depth is regressed at later stages. At inference time, $S3$ is the final depth output. ReLU \cite{nair2010rectified} and batch normalization \cite{ioffe2015batch} are adopted after each convolution in stacked hourglass and APC. 


We use groundtruth, $D_{gt}$, to directly supervise the regression and calculate loss terms for each stage output. The corresponding mean square error losses are computed as follows.
\begin{equation}
     L_{i}=\|D_{gt}-Si\|^2_2,  \forall i \in [1,3].
     \label{fused}
\end{equation}
The total loss is $L_1+L_2+L_3+L_c$. Note that $D_{gt}$ from KITTI Depth Completion does not contain points on the upper scenes. We find that using more stages or deeper network would cause overfitting on the lower parts and yield unstructured upper scenes.

\section{Experiments}

\vspace{-5pt}
\subsection{Sparse Depth Completion}
\vspace{-5pt}
\begin{figure*}[hbt!]
    \centering
    \includegraphics[width=1.0\linewidth]{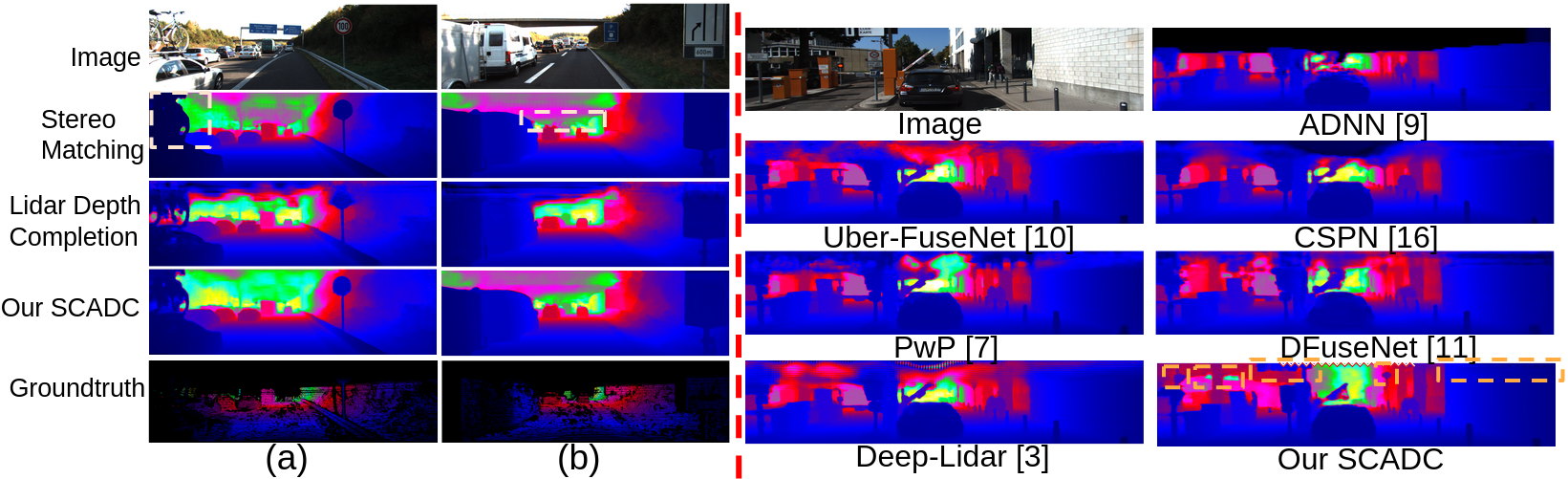}
    \vspace{-20pt}
    \caption{\textbf{(Left) Qualitative results of stereo matching (PSMNet \cite{chang2018pyramid}), SSDC (direct lidar completion \cite{ma2019self}), and our SCADC on KITTI Depth Completion validation set.} We show driving scenarios of large trucks beside and cars with loads. Vehicle structures extend to upper scenes. SSDC fails to regress upper structures. Shape distortion of PSMNet could be seen in highlights (a) Bicycle contour. (b) Bridge structure bleeds into the background and creates irregular estimations. \textbf{(Right) Comparison on KITTI Depth Completion test set.} Results of other works are directly from KITTI website. ADNN \cite{chodosh18} shows null on the upper since there are no groundtruth points. We are the only that reconstructs upper scene structures.}
    \label{all}
    \vspace{-12pt}
\end{figure*}

\textbf{Dataset.} We evaluate spare depth completion on KITTI Depth Completion Benchmark, which contains 42K stereo pairs and lidar scans as training data and 3.4K frames for validation. Following \cite{ma2019self}, we uniformly bottom crop the size to $352\times 1216$. Inputs to our SCADC are generated by PSMNet \cite{chang2018pyramid} and SSDC \cite{ma2019self}. We use their released code and best-pretrained weights on KITTI.

\textbf{Error Metrics.} We follow error metrics from previous works. (1) RMSE: root mean square error; (2) Rel: mean absolute relative error; (3) $\delta_i$: percentage of predicted pixels where the relative error is within 1.25$^i$. Formally,
\begin{equation}
    \vspace{-7pt}
     \delta_i = \frac{|\{\hat{d}:max(\frac{\hat{d}}{d},\frac{d}{\hat{d}})<1.25^i \}|} {|\{d\}|},
     \label{delat}
\end{equation}
where $|.|$ denotes the cardinality of a set. $\hat{d}$ and $d$ are prediction and groundtruth. Most studies adopt $i=1,2,3$.

\textbf{Results.} The quantitative comparison of depth error on KITTI Depth Completion val set is in Table \ref{table:input_comp}. Depth maps for comparison from PSMNet \cite{chang2018pyramid} and SSDC \cite{ma2019self} are generated following their steps. Note that the numerical results only evaluate depth estimations on the lower scenes. A qualitative comparison is shown in Fig.\ref{all} Left. From both numerical and visual results, although PSMNet produces more structured upper scenes than SSDC, the depth estimation error is larger on the lower part. By contrast, while SSDC has a smaller numerical error, it creates irregular and unstructured depth estimations on the upper scenes. Our SCADC combines the advantages of both stereo matching and depth completion to generate \textit{both scene completeness-aware and precise depth estimations.} We also compare with other depth completion methods on KITTI test set. The comparison is shown in Fig.\ref{all} Right. Our SCADC is the only work that successfully reconstructs the upper scene structures among the comparison.

\begin{table}[bt!]
\begin{center}
\caption{Evaluation on KITTI Depth Completion val set.}
\label{table:input_comp}
\footnotesize
\begin{tabular}{|p{1.2cm}<{\centering}||p{0.9cm}<{\centering}|p{0.9cm}<{\centering}|p{0.9cm}<{\centering}|p{0.5cm}<{\centering}|p{0.5cm}<{\centering}|p{0.5cm}<{\centering}|}
\hline
Methods & RMSE &Rel & $\delta$1 & $\delta$2 & $\delta$3\\
\hline
PSMNet & 2.4107 & 0.1296 & 98.6 & 99.8 & 99.9\\
SSDC  & 1.0438 & \textbf{0.0191} & 99.3 & 99.8 & 99.9\\
SCADC  & \textbf{1.0096} & 0.0226 & \textbf{99.5} & \textbf{99.9} & \textbf{100.0}\\
\hline
\end{tabular}
\end{center}
\vspace{-25pt}
\end{table}



\begin{table}[bt!]
\begin{center}
\caption{Comparison on KITTI Semantic Segmentation dataset. Our depth could enhance SSMA performance.}
\footnotesize
\label{table:semantic}
\begin{tabular}{|p{4.0cm}<{\centering}||p{1cm}<{\centering}|}
\hline
Methods & mIoU \\
\hline
SDNet \cite{ochs2019sdnet}& 51.15 \\
SGDepth \cite{klingner2020self} & 53.04 \\
SSMA \cite{valada19ijcv}& 54.76\\
SSMA + Our SCADC depth & \textbf{61.57} \\
\hline
\end{tabular}
\vspace{-20pt}
\end{center}
\end{table}

\begin{figure}[hbt!]
    \centering
    \includegraphics[width=1.0\linewidth]{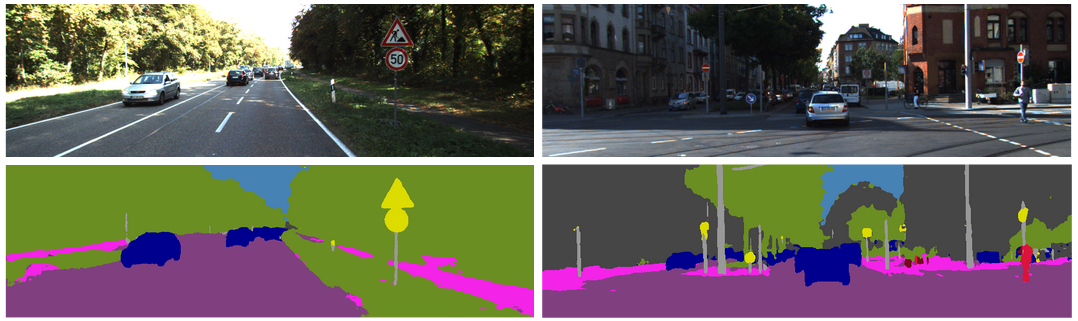}
    \vspace{-10pt}
    \caption{Semantic Segmentation results of SSMA with depth from our SCADC on KITTI Semantic Segmentation dataset.}
    \label{semantic}
    \vspace{-10pt}
\end{figure}

\vspace{-10pt}
\subsection{Outdoor RGBD Semantic Segmentation}
\vspace{-5pt}
\textbf{Datasets.} We next validate our SCADC on outdoor semantic segmentation. KITTI Semantic Segmentation dataset contains only images, and we match the corresponding lidar frames in KITTI Raw that contains all public raw data. We split the fetched data into 6:1 training and validation set. Although Cityscapes has more data for semantic segmentation, they only adopt stereo cameras as the depth acquisition.


\textbf{Evaluations.} SSMA \cite{valada19ijcv} is a high-performing framework on outdoor RGBD semantic segmentation. We follow SSMA's setting and use their Cityscapes pretrains to perform fine-tuning on KITTI. Standard mean intersection over union (mIoU) is adopted as the metrics. Two other RGBD outdoor semantic segmentation methods SDNet \cite{ochs2019sdnet} and SGDepth \cite{klingner2020self} are included for comparison. The quantitative and qualitative results are in Table \ref{table:semantic} and Fig. \ref{semantic}. From the results, our scene completeness-aware and precise depth could further help performance improvements. In the visual results, finer structures such as road signs and traffic poles that extend to the upper scenes could be clearly segmented.


\vspace{-9pt}
\section{CONCLUSIONS}
\vspace{-8pt}
Our SCADC combines the advantages of scene completeness from stereo matching to help lidar depth completion, obtaining both scene-complete and precise depth maps. Our APC module predicts confidences of lidars with a guiding supervision. With APC, information fusion from the two modalities is successfully performed. We show that our depth maps have good upper scene control for practical scenarios that objects of interest extend to the upper scenes. Furthermore, we validate that our depth maps help outdoor RGBD scene understanding, showing more values of depth completion.

\bibliographystyle{IEEEbib}
{\footnotesize
\bibliography{strings,refs}}

\begin{thebibliography}{10}

\bibitem{wu2023meta}
Cho-Ying Wu, Yiqi Zhong, Junying Wang, and Ulrich Neumann,
\newblock ``Meta-optimization for higher model generalizability in single-image
  depth prediction,''
\newblock {\em arXiv preprint arXiv:2305.07269}, 2023.

\bibitem{wu2020geometry}
Cho-Ying Wu, Xiaoyan Hu, Michael Happold, Qiangeng Xu, and Ulrich Neumann,
\newblock ``Geometry-aware instance segmentation with disparity maps,''
\newblock {\em arXiv preprint arXiv:2006.07802}, 2020.

\bibitem{wu2022toward}
Cho-Ying Wu, Jialiang Wang, Michael Hall, Ulrich Neumann, and Shuochen Su,
\newblock ``Toward practical monocular indoor depth estimation,''
\newblock in {\em Proceedings of the IEEE/CVF Conference on Computer Vision and
  Pattern Recognition}, 2022, pp. 3814--3824.

\bibitem{wu2023inspacetype}
Cho-Ying Wu, Quankai Gao, Chin-Cheng Hsu, Te-Lin Wu, Jing-Wen Chen, and Ulrich
  Neumann,
\newblock ``Inspacetype: Reconsider space type in indoor monocular depth
  estimation,''
\newblock {\em arXiv preprint arXiv:2309.13516}, 2023.

\bibitem{xu2020grid}
Qiangeng Xu, Xudong Sun, Cho-Ying Wu, Panqu Wang, and Ulrich Neumann,
\newblock ``Grid-gcn for fast and scalable point cloud learning,''
\newblock in {\em CVPR}, 2020, pp. 5661--5670.

\bibitem{uhrig2017sparsity}
Jonas Uhrig, Nick Schneider, Lukas Schneider, Uwe Franke, Thomas Brox, and
  Andreas Geiger,
\newblock ``Sparsity invariant cnns,''
\newblock in {\em IEEE International Conference on 3D Vision (3DV)}, 2017, pp.
  11--20.

\bibitem{Ma2017SparseToDense}
Fangchang Ma and Sertac Karaman,
\newblock ``Sparse-to-dense: Depth prediction from sparse depth samples and a
  single image,''
\newblock in {\em IEEE International Conference on Robotics and Automation
  (ICRA)}, 2018.

\bibitem{qiu2018deeplidar}
Jiaxiong Qiu, Zhaopeng Cui, Yinda Zhang, Xingdi Zhang, Shuaicheng Liu, Bing
  Zeng, and Marc Pollefeys,
\newblock ``Deeplidar: Deep surface normal guided depth prediction for outdoor
  scene from sparse lidar data and single color image,''
\newblock {\em IEEE Conference on Computer Vision and Pattern Recognition
  (CVPR)}, 2019.

\bibitem{ma2019self}
Fangchang Ma, Guilherme~Venturelli Cavalheiro, and Sertac Karaman,
\newblock ``Self-supervised sparse-to-dense: Self-supervised depth completion
  from lidar and monocular camera,''
\newblock in {\em 2019 International Conference on Robotics and Automation
  (ICRA)}. IEEE, 2019, pp. 3288--3295.

\bibitem{zhong2019deep}
Yiqi Zhong, Cho-Ying Wu, Suya You, and Ulrich Neumann,
\newblock ``Deep rgb-d canonical correlation analysis for sparse depth
  completion,''
\newblock in {\em Advances in Neural Information Processing Systems (NeurIPS)},
  2019, pp. 5332--5342.

\bibitem{leedepth}
Byeong-Uk Lee, Hae-Gon Jeon, Sunghoon Im, and In~So Kweon,
\newblock ``Depth completion with deep geometry and context guidance,''
\newblock in {\em IEEE International Conference on Robotics and Automation
  (ICRA)}, 2019.

\bibitem{wang2018pnp}
Tsun-Hsuan Wang, Fu-En Wang, Juan-Ting Lin, Yi-Hsuan Tsai, Wei-Chen Chiu, and
  Min Sun,
\newblock ``Plug-and-play: Improve depth estimation via sparse data
  propagation,''
\newblock in {\em IEEE International Conference on Robotics and Automation
  (ICRA)}, 2019.

\bibitem{imran2019depth}
Saif Imran, Yunfei Long, Xiaoming Liu, and Daniel Morris,
\newblock ``Depth coefficients for depth completion,''
\newblock in {\em CVPR}, 2019.

\bibitem{chodosh18}
Simon~Lucey Nathaniel~Chodosh, Chaoyang~Wang,
\newblock ``{Deep Convolutional Compressed Sensing for LiDAR Depth
  Completion},''
\newblock in {\em Asian Conference on Computer Vision (ACCV)}, 2018.

\bibitem{chen2019learning}
Yun Chen, Bin Yang, Ming Liang, and Raquel Urtasun,
\newblock ``Learning joint 2d-3d representations for depth completion,''
\newblock in {\em Proceedings of the IEEE International Conference on Computer
  Vision (ICCV)}, 2019, pp. 10023--10032.

\bibitem{shivakumar2019dfusenet}
Shreyas~S. Shivakumar, Ty~Nguyen, Steven~W. Chen, and Camillo~J. Taylor,
\newblock ``Dfusenet: Deep fusion of rgb and sparse depth information for image
  guided dense depth completion,''
\newblock {\em arXiv preprint arXiv:1902.00761}, 2019.

\bibitem{geiger2012we}
Andreas Geiger, Philip Lenz, and Raquel Urtasun,
\newblock ``Are we ready for autonomous driving? the kitti vision benchmark
  suite,''
\newblock in {\em IEEE Conference on Computer Vision and Pattern Recognition
  (CVPR)}, 2012, pp. 3354--3361.

\bibitem{chang2018pyramid}
Jia-Ren Chang and Yong-Sheng Chen,
\newblock ``Pyramid stereo matching network,''
\newblock in {\em CVPR}, 2018, pp. 5410--5418.

\bibitem{wang2014stereo}
Qiaosong Wang, Zhan Yu, Christopher Rasmussen, and Jingyi Yu,
\newblock ``Stereo vision--based depth of field rendering on a mobile device,''
\newblock {\em Journal of Electronic Imaging}, vol. 23, no. 2, pp. 023009,
  2014.

\bibitem{valada19ijcv}
Abhinav Valada, Rohit Mohan, and Wolfram Burgard,
\newblock ``Self-supervised model adaptation for multimodal semantic
  segmentation,''
\newblock {\em International Journal of Computer Vision (IJCV)}, jul 2019,
\newblock Special Issue: Deep Learning for Robotic Vision.

\bibitem{cheng2018learning}
Xinjing Cheng, Peng Wang, and Ruigang Yang,
\newblock ``Learning depth with convolutional spatial propagation network,''
\newblock {\em European Conference on Computer Vision (ECCV)}, 2018.

\bibitem{park2020non}
Jinsun Park, Kyungdon Joo, Zhe Hu, Chi-Kuei Liu, and In~So Kweon,
\newblock ``Non-local spatial propagation network for depth completion,''
\newblock {\em European Conference on Computer Vision (ECCV)}, 2020.

\bibitem{xu2019depth}
Yan Xu, Xinge Zhu, Jianping Shi, Guofeng Zhang, Hujun Bao, and Hongsheng Li,
\newblock ``Depth completion from sparse lidar data with depth-normal
  constraints,''
\newblock in {\em Proceedings of the IEEE International Conference on Computer
  Vision (ICCV)}, 2019, pp. 2811--2820.

\bibitem{hirschmuller2007stereo}
Heiko Hirschmuller,
\newblock ``Stereo processing by semiglobal matching and mutual information,''
\newblock {\em IEEE Transactions on pattern analysis and machine intelligence
  (TPAMI)}, vol. 30, no. 2, pp. 328--341, 2007.

\bibitem{hirschmuller2005accurate}
Heiko Hirschmuller,
\newblock ``Accurate and efficient stereo processing by semi-global matching
  and mutual information,''
\newblock in {\em IEEE Conference on Computer Vision and Pattern Recognition
  (CVPR)}. IEEE, 2005, vol.~2, pp. 807--814.

\bibitem{gehrig2009real}
Stefan~K Gehrig, Felix Eberli, and Thomas Meyer,
\newblock ``A real-time low-power stereo vision engine using semi-global
  matching,''
\newblock in {\em International Conference on Computer Vision Systems (ICCV)}.
  Springer, 2009, pp. 134--143.

\bibitem{hermann2012iterative}
Simon Hermann and Reinhard Klette,
\newblock ``Iterative semi-global matching for robust driver assistance
  systems,''
\newblock in {\em Asian Conference on Computer Vision (ACCV)}. Springer, 2012,
  pp. 465--478.

\bibitem{Guo_2019_CVPR}
Xiaoyang Guo, Kai Yang, Wukui Yang, Xiaogang Wang, and Hongsheng Li,
\newblock ``Group-wise correlation stereo network,''
\newblock in {\em Proceedings of the IEEE/CVF Conference on Computer Vision and
  Pattern Recognition (CVPR)}, June 2019.

\bibitem{Zhang_2019_CVPR}
Feihu Zhang, Victor Prisacariu, Ruigang Yang, and Philip~H.S. Torr,
\newblock ``Ga-net: Guided aggregation net for end-to-end stereo matching,''
\newblock in {\em Proceedings of the IEEE/CVF Conference on Computer Vision and
  Pattern Recognition (CVPR)}, June 2019.

\bibitem{forsyth2003modern}
David~A Forsyth and Jean Ponce,
\newblock ``Computer vision: A modern approach,''
\newblock 2003.

\bibitem{Jaritz_2018}
Maximilian Jaritz, Raoul~De Charette, Emilie Wirbel, Xavier Perrotton, and
  Fawzi Nashashibi,
\newblock ``Sparse and dense data with cnns: Depth completion and semantic
  segmentation,''
\newblock {\em IEEE International Conference on 3D Vision (3DV)}, 2018.

\bibitem{van2019sparse}
Wouter Van~Gansbeke, Davy Neven, Bert De~Brabandere, and Luc Van~Gool,
\newblock ``Sparse and noisy lidar completion with rgb guidance and
  uncertainty,''
\newblock in {\em 2019 16th International Conference on Machine Vision
  Applications (MVA)}. IEEE, 2019, pp. 1--6.

\bibitem{newell2016stacked}
Alejandro Newell, Kaiyu Yang, and Jia Deng,
\newblock ``Stacked hourglass networks for human pose estimation,''
\newblock in {\em European conference on computer vision (ECCV)}. Springer,
  2016, pp. 483--499.

\bibitem{nair2010rectified}
Vinod Nair and Geoffrey~E Hinton,
\newblock ``Rectified linear units improve restricted boltzmann machines,''
\newblock in {\em Proceedings of the 27th international conference on machine
  learning (ICML-10)}, 2010, pp. 807--814.

\bibitem{ioffe2015batch}
Sergey Ioffe and Christian Szegedy,
\newblock ``Batch normalization: Accelerating deep network training by reducing
  internal covariate shift,''
\newblock {\em arXiv preprint arXiv:1502.03167}, 2015.

\bibitem{ochs2019sdnet}
Matthias Ochs, Adrian Kretz, and Rudolf Mester,
\newblock ``Sdnet: Semantically guided depth estimation network,''
\newblock in {\em German Conference on Pattern Recognition (GCPR)}. Springer,
  2019, pp. 288--302.

\bibitem{klingner2020self}
Marvin Klingner, Jan-Aike Term{\"o}hlen, Jonas Mikolajczyk, and Tim
  Fingscheidt,
\newblock ``Self-supervised monocular depth estimation: Solving the dynamic
  object problem by semantic guidance,''
\newblock {\em European Conference on Computer Vision (ECCV)}, 2020.

\end{thebibliography}

\end{document}